# Updating velocities in heterogeneous comprehensive learning particle swarm optimization with low-discrepancy sequences

Yuelin Zhao, Feng Wu*, Jianhua Pang*, and Wanxie Zhong

*State Key Laboratory of Structural Analysis of Industrial Equipment, Department of Engineering Mechanics, Faculty of Vehicle Engineering and Mechanics, Dalian University of Technology, Dalian 116023, P.R.China; Guangdong Ocean University, Zhanjiang 524088, China; Shenzhen Institute of Guangdong Ocean University, Shenzhen 518120, China*

Email: zhaoyl9811@163.com, wufeng_chn@163.com, pangjianhua@gdou.edu.cn, wxzhong@dlut.edu.cn

Project supported by the National Natural Science Foundation of China grants (Nos. 11472076 and 51609034), the Science Foundation of Liaoning Province of China (No. 2021-MS-119), the Dalian Youth Science and Technology Star project (No. 2018RQ06), and the Fundamental Research Funds for the Central Universities grant (Nos. DUT20RC(5)009 and DUT20GJ216).

*Corresponding author: wufeng_chn@163.com, pangjianhua@gdou.edu.cn

September 2022

**Abstract:** Heterogeneous comprehensive learning particle swarm optimization (HCLPSO) is a type of evolutionary algorithm with enhanced exploration and exploitation capabilities. The low-discrepancy sequence (LDS) is more uniform in covering the search space than random sequences. In this paper, making use of the good uniformity of LDS to improve HCLPSO is researched. Numerical experiments are performed to show that it is impossible to effectively improve the search ability of HCLPSO by only using LDS to generate the initial population. However, if we properly choose some random sequences from HCLPSO's velocities updating formula and replace them with the deterministic LDS, we can obtain a more efficient algorithm. Compared with the original HCLPSO under the same accuracy requirement, the HCLPSO updating the velocities with the deterministic LDS can significantly reduce the iterations required for finding the optimal solution, without decreasing the success rate.

# 1. Introduction

To solve some real-life optimization problems that are discontinuous, non-convex and multimodal, many evolutionary algorithms have been developed, such as the particle swarm optimization (PSO) [1], differential evolution (DE) [2], [3], genetic algorithm (GA) [4], cuckoo search (CS) [5], grey wolf algorithm (GWO) [6]. These evolutionary algorithms have shown excellent performance in solving optimization problems in different fields [7]-[11]. Among the various evolutionary algorithms, PSO is widely used for its ease of implementation and rapid convergence to the optimum [12]-[15]. However, all the particles in PSO share the same global best information, which can cause the particles to cluster around the global best. If the global best is located near a local minimum, then the particles will easily fall into the local optimum, and the algorithm will suffer diversity loss near the local minimum [13]. Thus far, the improvements on the PSO-type algorithms can be divided into two aspects of their actions: (1) modifying the update formulas of position and/or velocity to balance the exploration behavior of global search and the exploitation nature of local search; (2) increasing the probability that the initial population covers the region of feasible solution in the search space.

In the first aspect, PSO has been greatly improved. In [16], a new parameter, called inertia weight, was first introduced into the original PSO to balance the exploration and exploitation capabilities. In [17], a set of constriction coefficients was introduced to control the convergence tendency of the particle swarm so that it includes both exploration and exploitation capabilities. Ref. [18] proposed a comprehensive learning particle swarm optimization algorithm (CLPSO), in which the particles have different degrees of exploration and exploitation abilities. Ref. [19] proposed a multi-swarm cooperative particle swarm optimization that employs one master swarm and several slave swarms to balance exploration and exploitation. In [20], dynamic multi-swarm particle swarm optimization (DMS-PSO) was proposed. DES-PSO divides the whole population into a large number of sub-swarms that are regrouped frequently by using various regrouping schedules and information is exchanged among the particles in the whole swarm. Recently, [21] proposed a modified CLPSO with two subpopulations, called heterogeneous CLPSO (HCLPSO). HCLPSO balances the exploration and exploitation capabilities of the particle swarm algorithms by comprehensively using adaptive control parameters, controlling information sharing between the particles, learning strategies and heterogeneous swarms, and it avoids the problems of the "oscillation phenomenon" [22] and the "two steps forward, one step back phenomenon" [23]. It has also become one of the recognized improved PSO algorithms with better performance.

In regard to the second aspect, most researchers focused on the improvement of the population initialization method. PSO and even most evolutionary algorithms are typically population-based stochastic search techniques, which share a common step: population initialization. The role of this step is to provide an initial solution. Then, the initial solution will be iteratively improved in the optimization process until the stop conditions are satisfied. However, due to the limited population size, the probability of covering the promising areas within the search space decreases by increasing the number of dimensions of the search space [24], [25]. It is usually assumed that the initial population uniformly covering the whole search space may help the algorithm find the optimal solution. Therefore, population initialization technology has been a widely studied issue in evolutionary algorithm research [26]. Simple random sampling is the most common method to initialize the population in the evolutionary algorithms [27]. However, the high discrepancy of the random initial population often leads to the degradation of the performance of the algorithm. Many researchers have considered uniformly distributed low-discrepancy sequences (LDS) to improve the algorithms. However, there seems to be no definite conclusion about the effect of the LDS on the performance of the evolutionary algorithm. Several studies claimed that the initial population generated by LDS could improve the performance of the evolutionary algorithm. Ref. [28] used the Halton sequence in a real-coded GA, and observed that the Halton sequence can improve the probability of finding the optimal solution of the GA and reduce the number of iterations required for optimization. Ref. [29] believed that the LDS can be used to improve the calculation accuracy and speed of PSO in high-dimensional space. Ref. [30] verified that the low-discrepancy initialization method can reduce the variation of the search results without losing accuracy and performance. However, some other researchers believe that the improvement of LDS on the evolutionary algorithm is not significant. By numerical experiments, Ref. [31] concluded that advanced population initialization methods have no significant impact on the performance of DE. The numerical results in [32] showed that random initialization is the best for PSO, exponential initialization is the best for CS, and DE and GA are not sensitive to initialization methods. Recently, [33] compared the effects of five famous population initialization methods on four evolutionary algorithms, including PSO, and concluded that all of the four evolutionary algorithms considered are not sensitive to initialization methods under sufficient iteration times.

Clearly, previous studies could not offer a definite conclusion on whether the influence of the LDS on evolutionary algorithms, especially PSO-type algorithms, is positive or negative. These unclear conclusions and contradictions motivated us to investigate this issue more deeply by using the PSO-type algorithms. A noteworthy observation is that the LDS causes the initial population to cover the entire search space much better. It was also observed in some references that the LDS can enhance the search ability of evolutionary algorithms in the early iterations [33]. These two observations at least show that the improvement of the coverage ability of the population can indeed promote an improvement of the search ability of the evolutionary algorithms. However, according to the research of Alaa Tharwat et al. [33], the performance of the initial population generated by the LDS is not superior to that generated by a random sequence. This makes us consider: why the improvement of algorithm search ability brought by the LDS cannot be maintained throughout the iteration process. What factor destroys the improvement of algorithm search ability brought by the LDS? We think it may be the randomness of evolutionary algorithms that destroys the coverage ability of the LDS. Taking the classical PSO algorithm as an example, the initial population generated by the LDS does have better coverage ability, which leads to a better search ability in the early iterations. However, the velocity updating mechanism of PSO in the iteration process depends on random sequences. As the iteration processes, randomness gradually accumulates, and the improvement in the population coverage ability brought by the LDS gradually declines. Therefore, we propose an assumption: If the random sequence in the velocity updating formula of PSO is replaced with the LDS, the population can always cover the space close to the optimal solution better for the entire iteration process, and finally the search ability of PSO can be significantly improved. The main focus of this report is to demonstrate the above assumption based on numerical experiments, which will be the basic work to improve the PSO-type algorithm. Considering that HCLPSO is one of the best PSO-type algorithms at present, we will carry out the research with HCLPSO.

Currently, LDSs can be divided into three types. The first type is based on number theory, such as the Hua Wang (HW) sequence [34]-[36], Halton sequence [37], Kronecker sequence [38], van der Corput sequence [39], Niederreiter sequence [40], Sobol sequence [41], and so on. The second type of LDS is based on an experimental design, such as the orthogonal array (OA) and the uniform experimental design [42], [43]. The experimental design method is a space filling algorithm that looks for uniformly dispersed points in a given range. Since its first introduction in 1980, experimental design methods have been widely used in industrial and computer simulation designs [44]. The third type of LDS is based on evolution algorithms. This type of LDS is generally generated by using the evolutionary algorithm to optimize the existing LDS. One of these types of sequences is the optimized Halton sequence (OHS) [45] which is optimized with a differential evolution. Another is the dynamics evolution sequence (DES) [46] which is generated by using the multi-body dynamic evolution model. The HWS, OA, OHS and DES selected from the above three types of LDSs will be used to demonstrate the proposed assumption. The research contributions of this paper can be summarized as follows:

（1） It will be shown again by numerical tests that the initial population generated by LDS will indeed produce a degree of advantage in the early iterations of HCLPSO. However, the advantage decreases iteratively because of the random behavior of HCLPSO.

（2） It was found that reasonable replacements of random sequences in the velocity updating formula of HCLPSO with LDSs can significantly improve the calculation performance in terms of computational times and number of iterations required for convergence.

（3） An open question is presented that: For general evolutionary algorithms, can the computational performance be significantly improved by appropriately reducing the randomness in its iterative scheme (and replacing the random sequence with LDS)?

The rest of this paper is organized as follows: Section 2 briefly reviews related work, including the iterative scheme of HCLPSO, and the LDSs used in numerical experiments. Section 3 discusses the effect of initial populations generated with different LDSs on the performance of HCLPSO. In Section 4, numerical experiments are provided to demonstrate that reasonably replacing the random sequences in the velocity updating with different LDSs can significantly improve the calculation performance of HCLPSO for both the low-dimensional and high-dimensional cases. Finally, concluding remarks and prospects for future work are provided in Section 5.

## 2. Related work

### 2.1 HCLPSO

PSO is a population-based evolutionary algorithm. The potential solution to a problem is treated as a flying bird, adjusting its position in the search space according to its own experience and that of other birds. In HCLPSO, to balance the exploration behavior of the global search and the exploitation nature of the local search, the swarm population is divided into two subpopulations, which are used for exploration and exploitation, respectively.

To facilitate our subsequent analysis, we represent HCLPSO in a matrix-vector form and denote the dimension of the problem as $D$, the number of populations as $N$, the maximum number of iterations as $G$, the objective function as $f$, and the feasible solution as $\boldsymbol{x} = (x_1, x_2, \cdots, x_D)^\mathrm{T} \in R^{D\times 1}$, where $x_i \in [a_i, b_i]$. Then let

$$\boldsymbol{a} = (a_1, a_2, \cdots, a_D)^\mathrm{T}, \quad \boldsymbol{b} = (b_1, b_2, \cdots, b_D)^\mathrm{T}. \tag{1}$$

Let the population be expressed as

$$X_g = \left[ x_{g,1}, x_{g,2}, \cdots, x_{g,N} \right], \tag{2}$$

then the initial population in HCLPSO is

$$X_0 = a \otimes 1_N + \varepsilon_0 \circ (b \otimes 1 - a \otimes 1), \tag{3}$$

where $\otimes$ denotes the Kronecker product, $\circ$ denotes the Hadamard product, $1_N$ is a $1 \times N$ vector of ones, and $\varepsilon_0$ is a $D \times N$ random matrix of numbers uniformly distributed between 0 and 1.

In each iteration of HCLPSO, the population is updated by

$$x_{g+1,i} = x_{g,i} + v_{g+1,i}, \quad i = 1, 2, \cdots, N \tag{4}$$

where $v_{g+1,i}$ represents the velocity of the $i$-th particle. In HCLPSO, the population is be classified into two types, i.e., the exploration and exploitation subpopulations, and the sizes of which are $N_1$ and $N_2$, respectively. The velocity updating formulas for these two types of subpopulations are different. For the exploration subpopulation, the velocity updating formula is

$$v_{g+1,i} = w_g v_{g,i} + k_g \varepsilon_{g,1,i} \circ (p_{g,i} - x_{g,i}), \quad 1 \leq i \leq N_1, \tag{5}$$

where $w_g = 0.99 - 0.79\, g/G$ and $k_g = 3 - 0.5\, g/G$ are used in the simulations and $\varepsilon_{g,1,i}$ is a $D \times 1$ vector of uniformly distributed random numbers between 0 and 1. $p_{g,i}$ is the comprehensive learning vector which is random, and is well-designed to make the $i$-th particle can learn from all the other particles' best experiences. Detailed information on the calculation of $p_{g,i}$ can be found in [21]. For the exploitation subpopulation, the velocity updating formula is

$$v_{g+1,i} = w_g v_{g,i} + c_{1g} \varepsilon_{g,2,i} \circ (p_{g,i} - x_{g,i}) + c_{2g} \varepsilon_{g,3,i} \circ (x_{g,\text{best}} - x_{g,i}), \quad N_1 < i \leq N_1 + N_2 \tag{6}$$

where $c_{1,g} = 2.5 - 2\, g/G$, $c_{2,g} = 0.5 + 2\, g/G$, $\varepsilon_{g,2,i}$ and $\varepsilon_{g,3,i}$ are two $D \times 1$ vectors of uniformly distributed randoms number between 0 and 1, and $x_{g,\text{best}} = \arg\min_{1 \leq i \leq N} \left( f(x_{g,i}) \right)$.

Many previous studies have tried to improve the performance of PSO by replacing the random matrix $\varepsilon_0$ in the initial population with a deterministic matrix generated with the LDS. However, it remains an open question as to whether the initial population generated using LDS can improve the performance of HCLPSO. In Section 3, this problem will be investigated numerically. In the rest of this section, four LDSs adopted in this research will be briefly introduced.

## 2.2 LDSs adopted in this report

### 2.2.1 Hua-Wang sequence (HWS)

The HW sequence [34-36] is defined as $\boldsymbol{P}_{\mathrm{HW}} = [\boldsymbol{x}_1, \boldsymbol{x}_2, \boldsymbol{x}_3, \cdots, \boldsymbol{x}_N]$, where

$$\boldsymbol{x}_i = \mathrm{f}(i \cdot \boldsymbol{\gamma}) \tag{7}$$

$$\boldsymbol{\gamma} = \left[ \mathrm{f}\left(2\cos\frac{2\pi}{p}\right), \mathrm{f}\left(2\cos\frac{4\pi}{p}\right), \cdots, \mathrm{f}\left(2\cos\frac{2\pi D}{p}\right) \right]^{\mathrm{T}}, \tag{8}$$

$p \geq 2D+3$ is a prime number, and $\mathrm{f}(x)$ represents is the fractional part of $x$.

### 2.2.2 Optimized Halton sequence (OHS)

The second LDS is the optimized Halton sequence proposed in [45]. It is generated by using a set of Halton generators optimized with the evolutionary algorithm. The optimized generators are available at http://vision.gel.ulaval.ca/~fmdrainville/permutations.html.

### 2.2.3 Dynamic evolution sequence (DES)

DES was a sequence recently developed in [46]. This sequence is based on the physical observation that the static solution of the multi-body problem is a kind of LDS. All particles in a given space move under the action of gravity and dynamic law. When all the particles are at rest, their coordinates form an LDS. The DESs with dimensions smaller than 20 can be downloaded from https://rocewea.com/2.html.

### 2.2.4 Orthogonal Array (OA)

OA is produced based on the experimental design method. It can provide combinations of positions that are evenly distributed, and is applied to generate the initial population of points that are scattered uniformly over the feasible solution space [42]. The computational cost of using an orthogonal array to construct large population points is often lower than that of using other uniform experimental design methods, and hence it was preferred by many researchers [47], [48].

# 3. Effect of population initializers on HCLPSO

In this section, numerical experiments are presented to test whether generating initial populations with LDSs can significantly improve the performance of HCLPSO. Although the effect of population initializers on HCLPSO was investigated in [33], the LDSs considered here are different from those in [33]. In particular, both the DMS and OHS are generated by optimizing the existing LDSs with the evolutionary algorithms.

## 3.1 Experimental settings

### 3.1.1 Measurement of performance

In this section, the HCLPSO with two different population initializers is considered as two different algorithms. To compare the performance of different algorithms, we used the following metrics:

Error tolerance ($\varepsilon_{tol}$): When the relative error between the fitness of the best solution and the true optimal solution is smaller than $\varepsilon_{tol}$, the algorithm is considered to be convergent.

Convergence speed (CS): This metric indicates how many iterations there are after which the algorithm converges. For two different algorithms, when the same $\varepsilon_{tol}$ is given, the fewer iterations required for the algorithm to converge, the faster the convergence speed of the algorithm and the more powerful the search ability of the algorithm. Due to the randomness existing in the iterations, 60 runs are performed for each test function and each algorithm. The average convergence curve of the 60 runs is recorded, and the CS for a given error tolerance is evaluated based on the average convergence curve. If the CS is greater than the maximum number of iterations $G$, it is claimed that the algorithm fails to find the global optimum under the given error tolerance.

### 3.1.2 Test functions and parameters

Here, 17 benchmark functions with $D=10$ selected from the 30 widely used single-objective benchmark functions provided by the CEC 2017 special session [49] are adopted. The name, the topology, and the true optimal solution ($Z^*$) of all the test functions are highlighted in Table 1. For for the other 13 functions, the original HCLPSO cannot find the optimal solutions under $\varepsilon_{tol} = 5\%$ and $G = 7500$, and they are discarded.

Table 1 Details of the benchmark functions

| Topology | No. | Function name | $Z^*$ |
|---|---|---|---|
| Unimodal | $F_1$ | Shifted and rotated Zakharov function | 300 |
| Simple multimodal Fns. | $F_2$ | Shifted and rotated Rosenbrock's function | 400 |
| | $F_3$ | Shifted and rotated Rastrigin's function | 500 |
| | $F_4$ | Shifted and rotated Expaned Schaffer F6 function | 600 |
| | $F_5$ | Shifted and rotated Lunacek Bi-Ratrigin's function | 700 |
| | $F_6$ | Shifted and rotated Non-continuous Rastrigin's function | 800 |
| | $F_7$ | Shifted and rotated Lvey function | 900 |
| Hybrid Fns. | $F_8$ | Zakharov, Rosenbrock, Rastrigin | 1100 |
| | $F_9$ | High-conditioned elliptic; Ackley; Schaffer; Rastrigin | 1400 |
| | $F_{10}$ | Bent Cigar; HGBat; Rastrigin; Rosenbrock | 1500 |
| | $F_{11}$ | Expanded schaffer; HGBat; Rosenbrock; Modified Schwefel; Rastrigin | 1600 |
| | $F_{12}$ | Katsuura; Ackely; Expanded Griewank plus Rosenbrock; Schwefel; Rastrigin | 1700 |
| | $F_{13}$ | Bent Cigar; Griewank plus Rosenbrock; Rastrigin; Expanded Schaffer | 1900 |
| | $F_{14}$ | Katsuura; Ackley; Rastrigin; Schaffer; Modified Schwefel | 2000 |
| Composite Fns. | $F_{15}$ | Rosenbrock; High-conditioned Elliptic; Rastrigin | 2100 |
| | $F_{16}$ | Griewank; Rastrigin; Modified schwefel | 2200 |
| | $F_{17}$ | Ackley; Griewank; Rastrigin; High-conditioned Elliptic | 2400 |

Fns. Functions

In our experiments, we used two different error tolerances, i.e., $\varepsilon_{tol} = 1\%$ and $5\%$. The sizes of the exploration and exploitation subpopulations are $N_1 = 15$ and $N_2 = 25$, respectively. Other parameters involved in HCLPSO are the same as those used in [21].

## 3.2 Statistical analysis

Suppose there are $k$ algorithms and $M$ test functions, and each algorithm is used to solve the $M$ test functions, and produces $M$ different results. To statistically compare the performance of these $k$ algorithms, one of the common methods is the modified Friedman test[50], which is briefly described here.

The modified Friedman test ranks all the algorithms in terms of the numerical results for each test function. The rank of the best performing algorithm is one, the second-best algorithm is two, and so on. In the case of ties, the average rank will be assigned. According to the rank of each algorithm, the modified Friedman statistics $\tau_F$ are calculated as follows:

$$\tau_F = \frac{(M-1)\chi_F^2}{M(k-1)-\chi_F^2}, \quad \chi_F^2 = \frac{12M}{k(k+1)}\left[\sum_i R_i^2 - \frac{k(k+1)^2}{4}\right] \quad (9)$$

where $R_i$ is the average rank of the $i$-th algorithm. The smaller $R_i$ is, the better the performance of the $i$-th algorithm is. The null hypothesis of the modified Friedman test states that all the algorithms are equivalent, and thus their ranks should be equal. If the statistics $\tau_F$ are greater than the critical value $\tau_c$, which can be obtained according to the $F$-distribution with the confidence level $\alpha$, the null hypothesis will be rejected. In this case, the Nemenyi test is used to justify algorithm is significantly superior to the other algorithms. This test states that the performance of the two algorithms is significantly different if the corresponding average ranks differ by at least a critical difference, defined as follows:

$$CD = q_\alpha \sqrt{\frac{k(k+1)}{6M}} \quad (10)$$

where $q_\alpha$ can be obtained in terms of the distribution with the confidence level $\alpha$. In all the experiments, the confidence level is set to be $\alpha = 0.05$.

## 3.3 Performance evaluation of HCLPSO with different population initialization methods

Table 2 shows the CSs and the ranks of different algorithms under the condition of $\varepsilon_{tol} = 5\%$ and 1%. According to (9), the values of $\tau_F$ for $\varepsilon_{tol} = 1\%$ and 5% are 0.2960 and 0.4907, respectively. Both values are smaller than the critical value $\tau_c = 2.537$, which means the null hypothesis should be accepted. Therefore, it can be concluded that, with the same error tolerance, there is no significant difference among the numbers of iterations required for the different algorithms to converge. According to (10), the critical rank difference is $CD = 1.137$. Using the Nemenyi test for a pairwise comparison, it is observed that the difference between the AvgRks. of any two initializers is smaller than $CD$, which further shows that there is no significant difference between the performance of the different initializers. As a result, we can conclude that adopting LDS to generate the initial population cannot significantly improve the computational performance of HCLPSO.

Table 2 CSs and ranks of the five algorithms at different tolerance errors

|  | $\varepsilon_{tol} = 5\%$ | | | | | $\varepsilon_{tol} = 1\%$ | | | | |
|---|---|---|---|---|---|---|---|---|---|---|
|  | Rand | DES | HWS | OHS | OA | Rand | DES | HWS | OHS | OA |
| $F_1$ | 2238(2) | 2247(3.5) | 2247(3.5) | 2253(5) | 2224(1) | 2410(5) | 2398(2) | 2404(4) | 2393(1) | 2402(3) |
| $F_2$ | 986(5) | 967(3) | 962(2) | 954(1) | 978(4) | 3299(5) | 3212(2) | 3252(3) | 3296(4) | 3180(1) |
| $F_3$ | 2006(3) | 1951(1) | 2042(4) | 1995(2) | 2053(5) | 4135(1) | 4335(2) | 4414(3) | 4569(4) | 4875(5) |
| $F_4$ | 84(5) | 76(1) | 78(3) | 77(2) | 80(4) | 1419(2.5) | 1432(5) | 1420(4) | 1397(1) | 1419(2.5) |
| $F_5$ | 2604(3) | 2643(5) | 2571(2) | 2624(4) | 2565(1) | -(3) | -(3) | -(3) | -(3) | -(3) |
| $F_6$ | 637(1) | 640(2) | 792(5) | 706(3) | 711(4) | 3096(2) | 3128(4) | 3200(5) | 3068(1) | 3109(3) |
| $F_7$ | 1296(5) | 1283(4) | 1269(2) | 1276(3) | 1252(1) | 1675(2) | 1682(3.5) | 1671(1) | 1688(5) | 1682(3.5) |
| $F_8$ | 1037(2) | 1073(5) | 1024(1) | 1049(4) | 1039(3) | 2006(5) | 2002(4) | 1949(1) | 2001(3) | 1973(2) |
| $F_9$ | 2155(5) | 2126(3.5) | 2126(3.5) | 2089(1) | 2102(2) | 4846(2) | 4598(1) | 4984(4) | 4857(3) | 5249(5) |
| $F_{10}$ | 2565(2) | 2563(1) | 2570(3) | 2605(4) | 2683(5) | 3870(2) | 3812(1) | 4353(4) | 4565(5) | 4212(3) |
| $F_{11}$ | 889(1) | 902(2) | 973(5) | 920(3) | 940(4) | 1621(1) | 1651(3) | 1680(4) | 1642(2) | 1682(5) |
| $F_{12}$ | 644(1) | 759(5) | 747(4) | 703(3) | 686(2) | 2712(1) | 2793(4) | 2769(3) | 2719(2) | 2801(5) |
| $F_{13}$ | 2279(4) | 2366(5) | 2109(2) | 2148(3) | 2080(1) | 3681(5) | 3380(4) | 3302(3) | 3250(2) | 3113(1) |
| $F_{14}$ | 537(2) | 586(4) | 563(3) | 536(1) | 630(5) | 2121(5) | 2095(2) | 2101(3) | 2082(1) | 2118(4) |
| $F_{15}$ | 7500(4) | 7500(4) | 7500(4) | 1797(2) | 1549(1) | -(3) | -(3) | -(3) | -(3) | -(3) |
| $F_{16}$ | 1471(2) | 1544(5) | 1499(4) | 1497(3) | 1299(1) | -(3) | -(3) | -(3) | -(3) | -(3) |
| $F_{17}$ | 3621(5) | 2511(3) | 1972(2) | 2716(4) | 1868(1) | -(3) | -(3) | -(3) | -(3) | -(3) |
| Avks. | 3.059(3) | 3.353(5) | 3.118(4) | 2.824(2) | 2.647(1) | 2.971(3) | 2.912(2) | 3.176(4) | 2.706(1) | 3.235(5) |

Avks. Average ranks

Although there is not a significant difference between the CSs of the different algorithms at $\varepsilon_{tol} = 5\%$ or $1\%$, the enhancement of the exploration ability in the early iterations when using LDS to generate the initial population is observed. Fig. 1 shows the convergence curves of HCLPSO with different initializers for different test functions. As observed, within the first 10 iterations, the convergence curves of the HCLPSO with LDS initializers decrease faster than those of the HCLPSO with random initializer, which shows that a better uniformity of the initial population can enhance the search ability to a certain extent. However, this enhancement is weakened with the proceeding of iterations. It is not clear which factor causes the gradual weakening. In this paper, it is believed that the randomness of HCLPSO, which gradually destroys the uniformity of the population, leads to a gradual weakening. If this explanation holds, the following deduction can be asserted: Reducing the randomness of HCLPSO can help to maintain the enhancements from the LDS initializer and improve the numerical performance. In the next section, numerical examples will be provided to validate this deduction and explanation.

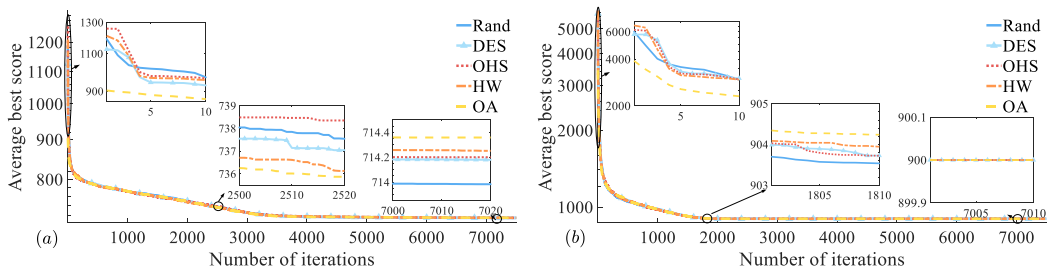

Fig. 1 The average convergence curves of different algorithms for: (a) $F_5$; and (b) $F_7$

## 4. Velocity updating with LDS

This section discusses whether it is the randomness of HCLPSO that weakens the enhancement caused by the LDS initializer, and whether reducing the randomness of HCLPSO is helpful in maintaining the enhancement from the LDS initializer and improving the numerical performance. In terms of (5) and (6), in addition to the comprehensive learning vector $p_{g,i}$ being random, $\varepsilon_{g,1}$, $\varepsilon_{g,2}$ and $\varepsilon_{g,3}$ are also random. The comprehensive learning vector is well-designed to help the particle select its own or another's historical best position to learn based on random numbers and a given learning probability. If the random numbers in $p_{g,i}$ are replaced with deterministic numbers, the learning objective of each particle will no longer change during the iterative process, which will inevitably result in the loss of the diversity of the population. Hence, we try to replace the random sequences $\varepsilon_{g,1}$, $\varepsilon_{g,2}$ and $\varepsilon_{g,3}$ with deterministic LDSs to reduce the randomness of HCLPSO and test the influence of this replacement on HCLPSO.

## 4.1 Comparison between the random number and LDS in velocity updating

The influence of replacing $\varepsilon_{g,1}$, $\varepsilon_{g,2}$ and $\varepsilon_{g,3}$ with deterministic LDSs on HCLPSO is tested numerically in this subsection. As $\varepsilon_{g,1}$ and $\varepsilon_{g,2}$ are used to control how much the particle learns from $p_{g,i}$, they are considered the same type of random factor, and $\varepsilon_{g,3}$ is another type of random factor. To test which random factors affect HCLPSO more, we provide three schemes. The first one, denoted by $HCLPSO_1$, replaces $\varepsilon_{g,1}$ and $\varepsilon_{g,2}$ with LDSs, and $\varepsilon_{g,3}$ is still random; the second one, denoted by $HCLPSO_2$, replaces $\varepsilon_{g,3}$ with the LDS, and $\varepsilon_{g,1}$ and $\varepsilon_{g,2}$ are still random; and the third one, denoted by $HCLPSO_0$, replaces all the $\varepsilon_{g,1}$, $\varepsilon_{g,2}$ and $\varepsilon_{g,3}$ with LDSs. The original HCLPSO, denoted by $HCLPSO_3$, is considered for comparison. The LDS used here is the OHS. The other parameters used are the same as those in Subsection 3.1.

Table 3 CSs and ranks of the four algorithms at different tolerance errors

| Fns. | $\varepsilon_{tol} = 5\%$ | | | | $\varepsilon_{tol} = 1\%$ | | | |
|---|---|---|---|---|---|---|---|---|
| | $HCLPSO_0$ | $HCLPSO_1$ | $HCLPSO_2$ | $HCLPSO_3$ | $HCLPSO_0$ | $HCLPSO_1$ | $HCLPSO_2$ | $HCLPSO_3$ |
| $F_1$ | 2261 (4) | 949 (2) | 2008 (3) | 840 (1) | 2405 (4) | 1065 (2) | 2139 (3) | 941 (1) |
| $F_2$ | 964 (4) | 279 (2) | 930 (3) | 269 (1) | 3301 (3) | 1750 (1) | 3717 (4) | 2173 (2) |
| $F_3$ | 1983 (4) | 827 (2) | 1806 (3) | 782 (1) | 4160 (4) | 2383 (1) | 3344 (3) | 3262 (2) |
| $F_4$ | 79 (2) | 67 (1) | 86 (4) | 83 (3) | 1414 (4) | 410 (2) | 1268 (3) | 361 (1) |
| $F_5$ | 2611 (4) | 1092 (2) | 2291 (3) | 1033 (1) | - (2.5) | - (2.5) | - (2.5) | -(2.5) |
| $F_6$ | 635 (3) | 366 (2) | 659 (4) | 335 (1) | 3201 (4) | 1825 (1) | 2971 (3) | 2155 (2) |
| $F_7$ | 1293 (4) | 351 (2) | 1136 (3) | 306 (1) | 1656 (4) | 517 (2) | 1521 (3) | 444 (1) |
| $F_8$ | 1059 (4) | 316 (2) | 987 (3) | 289 (1) | 1987 (4) | 713 (2) | 1784 (3) | 633 (1) |
| $F_9$ | 2097 (4) | 1174 (1) | 1964 (3) | 1195 (2) | 4753 (3) | 4287 (2) | 4181 (1) | - (4) |
| $F_{10}$ | 2587 (4) | 1523 (2) | 2386 (3) | 1421 (1) | 4073 (3) | 3047 (1) | 3718 (2) | - (4) |
| $F_{11}$ | 935 (4) | 393 (2) | 905 (3) | 377 (1) | 1654 (4) | 711 (1) | 1546 (3) | 764 (2) |
| $F_{12}$ | 687 (3) | 343 (2) | 745 (4) | 340 (1) | 2826 (4) | 1745 (1) | 2769 (3) | 2278 (2) |
| $F_{13}$ | 1988 (3) | 1117 (1) | 2089 (4) | 1310 (2) | 3127 (3) | 2258 (1) | 3039 (2) | Inf (4) |
| $F_{14}$ | 598 (4) | 308 (2) | 492 (3) | 288 (1) | 2057 (4) | 984 (1) | 1941 (3) | 1123 (2) |
| $F_{15}$ | 2588 (4) | 1777 (3) | 1491 (2) | 1407 (1) | -(2.5) | - (2.5) | -(2.5) | - (2.5) |
| $F_{16}$ | 1410 (4) | 475 (2) | 1275 (3) | 424 (1) | - (2.5) | - (2.5) | - (2.5) | - (2.5) |
| $F_{17}$ | 3153 (1) | 3391 (2) | 6745 (3) | - (4) | - (2.5) | - (2.5) | - (2.5) | - (2.5) |
| Avgk. | 3.529(4) | 1.882(2) | 3.176(3) | 1.412(1) | 3.412(4) | 1.647(1) | 2.706(3) | 2.235(2) |

The CSs and the Friedman ranks of the different algorithms under $\varepsilon_{tol} = 5\%$ and 1% are listed in Table 3, where "-" indicates that the algorithm cannot converge to the given error tolerance. As shown in Table 3, the AvgRks. of $HCLPSO_1$, $HCLPSO_2$ and $HCLPSO_3$ are smaller than that of $HCLPSO_0$ for both $\varepsilon_{tol} = 5\%$ and $\varepsilon_{tol} = 1\%$. When $\varepsilon_{tol} = 5\%$, $HCLPSO_3$ performs best, and when $\varepsilon_{tol} = 1\%$, $HCLPSO_1$ is the best. According to (9), the modified Friedman statistics $\tau_F$ for $\varepsilon_{tol} = 1\%$ and 5% are 25.733 and 8.033, respectively. Both values are greater than the critical value $\tau_c = 2.537$, which indicates that the performance of the four algorithms is significantly different. Using the Nemenyi test for pairwise comparisons, the difference between the AvgRks. of $HCLPSO_1$ and $HCLPSO_0$ is greater than the critical difference $CD = 1.137$ for both cases of $\varepsilon_{tol} = 5\%$ and 1%. The same is true for the difference between the AvgRks. of $HCLPSO_3$ and $HCLPSO_0$. The difference between the AvgRks. of $HCLPSO_2$ and $HCLPSO_0$ is smaller than the critical difference. According to the Nemenyi test, $HCLPSO_1$ and $HCLPSO_3$ perform much better than $HCLPSO_0$. As a result, it could be concluded that properly reducing the randomness in the velocity updating of HCLPSO can improve the numerical performance. The influence of $\varepsilon_{g,1}$ and $\varepsilon_{g,2}$ on HCLPSO is more remarkable than that of $\varepsilon_{g,3}$. Given the same error tolerance, the iterations required for convergence can be greatly decreased by replacing $\varepsilon_{g,1}$ and $\varepsilon_{g,2}$ in the velocity updating with the deterministic LDS. Since the AvgRks. of $HCLPSO_1$ is smallest at $\varepsilon_{tol} = 1\%$, it will be used in the following numerical experiments

## 4.2 Influence of different LDSs in velocities updating

The influence of replacing $\varepsilon_{g,1}$ and $\varepsilon_{g,2}$ with different types of LDSs, i.e., DES, HWS, OHS, and OA, on HCLPSO is discussed in this subsection. All the parameters used in this subsection are the same as those in Section 3. The original HCLPSO is considered for comparison. The HCLPSO that updates velocities by using a different LDS is regarded as a different algorithm, and is referred to by the LDS in use. The original HCLPSO uses random numbers to update the velocities, and hence is called Rand.

Table 4 CSs and ranks of the five algorithms at different error tolerances

| Fns. | $\varepsilon_{tol} = 5\%$ | | | | | $\varepsilon_{tol} = 1\%$ | | | | |
|---|---|---|---|---|---|---|---|---|---|---|
| | Rand | DES | HWS | OHS | OA | Rand | DES | HWS | OHS | OA |
| $F_1$ | 2261(5) | 935(2) | 962(4) | 949(3) | 796(1) | 2405(5) | 1063(2) | 1093(4) | 1065(3) | 919(1) |
| $F_2$ | 964(5) | 256(2) | 263(3) | 279(4) | 201(1) | 3301(5) | 2011(4) | 1805(3) | 1750(2) | 1500(1) |
| $F_3$ | 1983(5) | 785(2) | 822(3) | 827(4) | 685(1) | 4160(5) | 2557(2) | 2652(3) | 2383(1) | 3427(4) |
| $F_4$ | 79(5) | 61(1) | 63(2) | 67(3) | 72(4) | 1414(5) | 379(2) | 404(3) | 410(4) | 329(1) |
| $F_5$ | 2611(5) | 1118(3) | 1156(4) | 1092(2) | 870(1) | -(3) | -(3) | -(3) | -(3) | -(3) |
| $F_6$ | 635(5) | 339(2) | 326(1) | 366(4) | 348(3) | 3201(5) | 1718(1) | 1797(2) | 1825(3) | 1923(4) |
| $F_7$ | 1293(5) | 335(2) | 339(3) | 351(4) | 277(1) | 1656(5) | 501(2) | 514(3) | 517(4) | 419(1) |
| $F_8$ | 1059(5) | 321(3) | 323(4) | 316(2) | 294(1) | 1987(5) | 721(3) | 747(4) | 713(2) | 651(1) |
| $F_9$ | 2083(5) | 1114(1) | 1115(2) | 1174(4) | 1146(3) | 5045(4) | 5018(3) | 3518(1) | 4287(2) | -(5) |
| $F_{10}$ | 2587(5) | 1484(1) | 1631(3) | 1523(2) | 1876(4) | 4073(4) | 2735(1) | 3270(3) | 3047(2) | -(5) |
| $F_{11}$ | 935(5) | 385(1) | 483(3) | 393(2) | 511(4) | 1654(5) | 694(1) | 865(3) | 711(2) | 925(4) |
| $F_{12}$ | 687(5) | 370(2) | 414(4) | 343(1) | 385(3) | 2826(5) | 1801(2) | 1962(3) | 1745(1) | 2152(4) |
| $F_{13}$ | 1988(5) | 1173(4) | 915(1) | 1117(3) | 949(2) | 3127(5) | 2434(4) | 1820(1) | 2258(3) | 1830(2) |
| $F_{14}$ | 623(5) | 306(2) | 299(1) | 308(3) | 356(4) | 2057(5) | 966(1) | 974(2) | 984(3) | 1002(4) |
| $F_{15}$ | -(4) | -(4) | -(4) | 1777(2) | 996(1) | -(3) | -(3) | -(3) | -(3) | -(3) |
| $F_{16}$ | 1387(5) | 363(2) | 468(3) | 475(4) | 231(1) | -(3) | -(3) | -(3) | -(3) | -(3) |
| $F_{17}$ | 3153(4) | 1338(2) | 1343(3) | 3391(5) | 500(1) | -(3) | -(3) | -(3) | -(3) | -(3) |
| Avgks. | 4.882(5) | 2.118(1) | 2.824(3) | 3.059(4) | 2.118(1) | 4.412(5) | 2.353(1) | 2.765(3) | 2.588(2) | 2.882(4) |

The CSs and the Friedman ranks of the different algorithms at the different error tolerances are listed in Table 4. According to (9), the modified Friedman statistics $\tau_F$ for $\varepsilon_{tol} = 1\%$ and 5% are 16.8876 and 5.7702, respectively. Both values are greater than the critical value $\tau_c = 2.827$, which shows the significant difference between the five algorithms. Using the Nemenyi test for pairwise comparisons, for both $\varepsilon_{tol} = 1\%$ and 5%, the difference between the AvgRks. of the HCLPSO with any of the considered four LDSs and the AvgRks. of the original HCLPSO is greater than the critical difference $CD = 1.4795$. Hence, given the same error tolerance, updating the velocities with any of the four considered LDSs can significantly reduce the number of iterations required for convergence. Among the four considered LDSs, the DES has the smallest AvgRks., and hence shows the most remarkable effect in improving the convergence speed.

Table 5 Computation times (s) and ranks of the five algorithms at different tolerance errors

| Fns. | $\varepsilon_{tol} = 5\%$ | | | | | $\varepsilon_{tol} = 1\%$ | | | | |
|---|---|---|---|---|---|---|---|---|---|---|
| | Rand | DES | HWS | OHS | OA | Rand | DES | HWS | OHS | OA |
| $F_1$ | 0.932(5) | 0.372(2) | 0.405(4) | 0.387(3) | 0.334(1) | 1.112(5) | 0.461(2) | 0.482(4) | 0.481(3) | 0.406(1) |
| $F_2$ | 0.474(5) | 0.110(2) | 0.116(3) | 0.117(4) | 0.092(1) | 1.531(5) | 0.904(4) | 0.849(3) | 0.839(2) | 0.740(1) |
| $F_3$ | 0.919(5) | 0.371(3) | 0.366(2) | 0.387(4) | 0.324(1) | 2.204(5) | 1.369(2) | 1.289(1) | 1.406(3) | 1.827(4) |
| $F_4$ | 0.050(5) | 0.038(2) | 0.032(1) | 0.038(3) | 0.043(4) | 0.730(5) | 0.207(2) | 0.221(4) | 0.211(3) | 0.175(1) |
| $F_5$ | 1.312(5) | 0.514(2) | 0.543(3) | 0.543(4) | 0.431(1) | -(3) | -(3) | -(3) | -(3) | -(3) |
| $F_6$ | 0.376(5) | 0.159(2) | 0.153(1) | 0.178(3) | 0.178(4) | 1.518(5) | 0.862(1) | 0.948(3) | 0.872(2) | 1.120(4) |
| $F_7$ | 0.707(5) | 0.184(2) | 0.203(4) | 0.197(3) | 0.143(1) | 0.890(5) | 0.256(2) | 0.278(4) | 0.271(3) | 0.211(1) |
| $F_8$ | 0.473(5) | 0.150(4) | 0.140(3) | 0.135(1) | 0.136(2) | 0.941(5) | 0.336(2) | 0.346(4) | 0.338(3) | 0.306(1) |
| $F_9$ | 1.111(5) | 0.562(2) | 0.575(3) | 0.599(4) | 0.538(1) | 2.766(4) | 2.314(3) | 2.013(1) | 2.148(2) | -(5) |
| $F_{10}$ | 1.213(5) | 0.658(1) | 0.752(2) | 0.884(3) | 1.089(4) | 2.044(4) | 1.492(1) | 1.522(2) | 1.664(3) | -(5) |
| $F_{11}$ | 0.422(5) | 0.163(2) | 0.263(4) | 0.163(1) | 0.259(3) | 0.770(5) | 0.316(1) | 0.388(4) | 0.357(2) | 0.363(3) |
| $F_{12}$ | 0.344(5) | 0.190(2) | 0.229(4) | 0.172(1) | 0.229(3) | 1.670(5) | 1.176(3) | 1.124(2) | 1.064(1) | 1.625(4) |
| $F_{13}$ | 1.469(5) | 0.877(3) | 0.749(1) | 1.042(4) | 0.770(2) | 2.982(5) | 2.655(4) | 1.988(1) | 2.414(3) | 2.011(2) |
| $F_{14}$ | 0.289(5) | 0.163(3) | 0.154(2) | 0.151(1) | 0.196(4) | 1.588(5) | 0.777(3) | 0.684(1) | 0.685(2) | 1.438(4) |
| $F_{15}$ | -(3) | -(3) | -(3) | -(3) | -(3) | -(3) | -(3) | -(3) | -(3) | -(3) |
| $F_{16}$ | 0.961(5) | 0.283(2) | 0.320(3) | 0.334(4) | 0.243(1) | -(3) | -(3) | -(3) | -(3) | -(3) |
| $F_{17}$ | 1.866(5) | 0.639(3) | 0.571(2) | 1.186(4) | 0.261(1) | -(3) | -(3) | -(3) | -(3) | -(3) |
| Avgks. | 4.882(5) | 2.353(2) | 2.647(3) | 2.941(4) | 2.176(1) | 4.412(5) | 2.471(1) | 2.706(3) | 2.588(2) | 2.824(4) |

To compare the computational efficiency of the five algorithms more clearly, we also calculate the average computation times required for the five algorithms to converge to $\varepsilon_{tol} = 1\%$ and 5%. The computation times and ranks of the different test functions are compared in Table 5. The $\tau_F$ for $\varepsilon_{tol} = 1\%$ and 5% are 14.5821 and 5.5070, respectively. Both values are greater than the critical value $\tau_c = 2.537$, which indicates significant differences among the computation times of the different algorithms. Furthermore, for both $\varepsilon_{tol} = 5\%$ and $\varepsilon_{tol} = 1\%$, the difference between the AvgRks. of the HCLPSO that uses any of the considered four LDSs and the AvgRks. of the original HCLPSO is greater than the critical difference $CD = 1.4795$, which again reflects the advantage of LDSs.

Table 6 NoSs and ranks of the five algorithms with different error tolerances

| Fns. | $\varepsilon_{tol} = 5\%$ | | | | | $\varepsilon_{tol} = 1\%$ | | | | |
|---|---|---|---|---|---|---|---|---|---|---|
| | Rand | DES | HWS | OHS | OA | Rand | DES | HWS | OHS | OA |
| $F_1$ | 60(3) | 60(3) | 60(3) | 60(3) | 60(3) | 60(3) | 60(3) | 60(3) | 60(3) | 60(3) |
| $F_2$ | 60(3) | 60(3) | 60(3) | 60(3) | 60(3) | 60(3) | 60(3) | 60(3) | 60(3) | 60(3) |
| $F_3$ | 60(3) | 60(3) | 60(3) | 60(3) | 60(3) | 56(3.5) | 57(2) | 59(1) | 56(3.5) | 53(5) |
| $F_4$ | 60(3) | 60(3) | 60(3) | 60(3) | 60(3) | 60(3) | 60(3) | 60(3) | 60(3) | 60(3) |
| $F_5$ | 60(3) | 60(3) | 60(3) | 60(3) | 60(3) | 0(3) | 0(3) | 0(3) | 0(3) | 0(3) |
| $F_6$ | 60(3) | 60(3) | 60(3) | 60(3) | 60(3) | 60(3) | 60(3) | 60(3) | 60(3) | 60(3) |
| $F_7$ | 60(3) | 60(3) | 60(3) | 60(3) | 60(3) | 60(3) | 60(3) | 60(3) | 60(3) | 60(3) |
| $F_8$ | 60(3) | 60(3) | 60(3) | 60(3) | 60(3) | 60(3) | 60(3) | 60(3) | 60(3) | 60(3) |
| $F_9$ | 60(2.5) | 59(5) | 60(2.5) | 60(2.5) | 60(2.5) | 43(3) | 38(4) | 48(1) | 44(2) | 31(5) |
| $F_{10}$ | 60(1.5) | 60(1.5) | 57(4) | 59(3) | 55(5) | 51(3) | 55(1) | 48(4) | 52(2) | 35(5) |
| $F_{11}$ | 60(3) | 60(3) | 60(3) | 60(3) | 60(3) | 60(3) | 60(3) | 60(3) | 60(3) | 60(3) |
| $F_{12}$ | 60(3) | 60(3) | 60(3) | 60(3) | 60(3) | 60(1.5) | 60(1.5) | 59(3.5) | 59(3.5) | 55(5) |
| $F_{13}$ | 60(2.5) | 60(2.5) | 60(2.5) | 58(5) | 60(2.5) | 59(1) | 51(5) | 57(3) | 53(4) | 58(2) |
| $F_{14}$ | 60(3) | 60(3) | 60(3) | 60(3) | 60(3) | 60(2.5) | 60(2.5) | 60(2.5) | 60(2.5) | 56(5) |
| $F_{15}$ | 55(3) | 50(4) | 48(5) | 56(1.5) | 56(1.5) | 0(3) | 0(3) | 0(3) | 0(3) | 0(3) |
| $F_{16}$ | 60(3) | 60(3) | 60(3) | 60(3) | 60(3) | 5(5) | 22(1) | 14(3) | 11(4) | 16(2) |
| $F_{17}$ | 52(4) | 60(1.5) | 56(3) | 50(5) | 60(1.5) | 0(4) | 0(4) | 2(2) | 3(1) | 0(4) |
| Avgks. | 2.912(2) | 2.970(3) | 3.117(4) | 3.117(4) | 2.882(1) | 2.970(4) | 2.823(2) | 2.765(1) | 2.912(3) | 3.529(5) |

To verify that the proposed method can substantially improve the convergence speed without reducing the original search ability of HCLPSO, we introduce a success rate index, the number of successes (NoS). NoS represents the number of times that the algorithm finds the approximate optima with a relative error less than $\varepsilon_{tol}$ in 60 runs. The NoSs and the corresponding Friedman ranks of the different algorithms are compared in Table 6. According to Table 6, the $\tau_F$ for $\varepsilon_{tol} = 1\%$ and 5% are 0.0807 and 0.6241, respectively. Both values are smaller than the critical value $\tau_c = 2.827$, indicating that there is no significant difference between the five algorithms. Therefore, given the same error tolerance, updating the velocities with any of the four considered LDSs will not reduce the success rate of finding the global optima.

According to Tables 4, 5 and 6, it can be concluded that updating the velocities with any of the four LDSs considered can significantly reduce the iterations and computation times required for HCLPSO to converge to the given error tolerance without decreasing the success rate. For $\varepsilon_{tol} = 5\%$, the velocity updating with OA ranks first in convergence speed and computation time. For $\varepsilon_{tol} = 1\%$, DES performs best, followed by OHS, HW and OA. Therefore, OA is recommended for low precision requirement problems, and DES is recommended for high precision requirement problems.

## 4.3 Performance of updating velocities with LDS in high dimensional space

This subsection evaluates the performance of updating the velocity using LDS in high dimensional space. The numbers of dimensions are 30 and 50. As shown in Tables 4 and 5, DES and OHS perform better at $\varepsilon_{tol} = 1\%$ than the other two LDSs. These two LDSs are therefore used in this subsection. Since the original HCLPSO cannot converge to $\varepsilon_{tol} = 5\%$ for some test functions in Table 1 in high dimensional space, $\varepsilon_{tol} = 10\%$ and $\varepsilon_{tol} = 20\%$ are used.

### 4.4.1 $D$ = 30

The performance of updating velocities with LDS in 30 dimensional space is first investigated. The experimental parameters, with the exception of $D = 30$, are the same as those in Section 3. For $F_9$ and $F_{11}$, all the considered algorithms fail to find the average optimal solutions with relative errors smaller than $\varepsilon_{tol} = 20\%$ in $G = 7500$ iterations, so they are discarded..

Table 7 CSs and ranks of the three algorithms at different tolerance errors

| Fns. | $\varepsilon_{tol} = 20\%$ | | | $\varepsilon_{tol} = 10\%$ | | |
|---|---|---|---|---|---|---|
| | Rand | DES | OHS | Rand | DES | OHS |
| $F_1$ | 5275(1) | 6402(3) | 6287(2) | 5507(1) | 6765(2) | 6581(1) |
| $F_2$ | 7118(3) | 5348(2) | 4652(1) | -(2) | -(2) | -(2) |
| $F_3$ | 2758(3) | 1823(2) | 1369(1) | 3414(3) | 2856(2) | 2021(1) |
| $F_4$ | 10(3) | 4(1) | 7(2) | 217(3) | 126(1) | 146(2) |
| $F_5$ | 3064(3) | 2179(2) | 1586(1) | -(2) | -(2) | -(2) |
| $F_6$ | 2269(3) | 1277(2) | 923(1) | 2963(3) | 2169(2) | 1464(1) |
| $F_7$ | 2296(3) | 989(2) | 819(1) | 2521(3) | 1131(2) | 942(1) |
| $F_8$ | 2120(3) | 980(2) | 858(1) | 2611(3) | 1506(2) | 1308(1) |
| $F_{10}$ | 7415(2) | 4870(2) | 4700(1) | -(2) | -(2) | -(2) |
| $F_{12}$ | 1967(3) | 899(2) | 865(2) | 2833(3) | 1700(2) | 1654(2) |
| $F_{13}$ | 5313(3) | 3816(2) | 3623(1) | -(3) | 4760(1) | 6313(2) |
| $F_{14}$ | 1938(3) | 1003(2) | 873(1) | 2830(3) | 2718(2) | 2062(1) |
| $F_{15}$ | 1352(3) | 551(2) | 453(1) | -(2) | -(2) | -(2) |
| $F_{16}$ | 1717(3) | 514(2) | 467(1) | 1998(3) | 772(2) | 669(1) |
| $F_{17}$ | 3082(3) | 1748(2) | 1561(1) | -(2) | -(2) | -(2) |
| Avgks. | 2.867(3) | 2(2) | 1.133(1) | 2.533(3) | 1.933(2) | 1.533(1) |

The CSs and Friedman ranks of the different algorithms with different error tolerances in 30 dimensional space are listed in Table 7. The Friedman critical value is $\tau_c = 3.340$, and the Nemenyi critical difference is $CD = 0.8559$. According to the results in Table 7, the $\tau_F$ for $\varepsilon_{tol} = 20\%$ and 10% are 17.1881 and 3.8977, respectively. It is obvious that there is a significant difference among the three algorithms considered. Both the AvgRks. of OHS and DES are smaller than that of Rand. When $\varepsilon_{tol} = 20\%$, both OHS and DES perform significantly better than Rand, and when $\varepsilon_{tol} = 10\%$, OHS remains significantly superior to Rand. For the test function $F_{13}$, DES and OHS successfully find the optimal solution under the accuracy requirement $\varepsilon_{tol} = 10\%$, while Rand fails.

To more intuitively compare the numerical performance of different algorithms, we plot the average convergence curves of the different algorithms corresponding to different test functions. The average convergence curves corresponding to $F_7$ and $F_{10}$ are displaced in Fig. 2, and the others are displaced in Supplementary Fig. 1. As shown, DES and OHS converge much faster than Rand in 30 dimensional space. The convergence speeds of DES and OHS are similar.

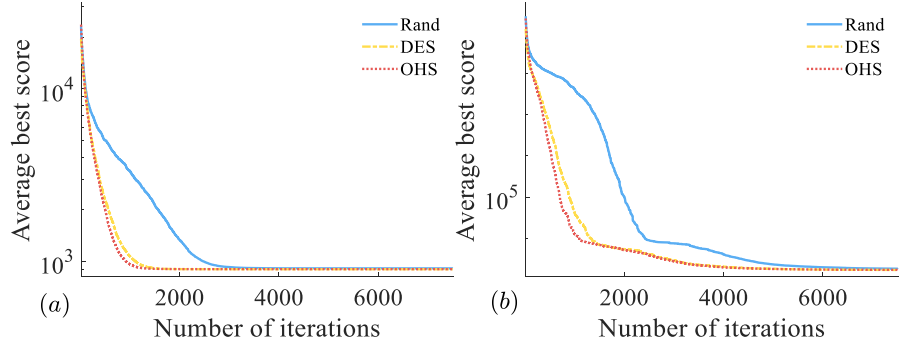

Fig. 2 The average convergence curves of the three algorithms with:(a) $F_7$; and (b) $F_{10}$

The NoSs of the different algorithms with different error tolerances are listed in Supplementary Table 1, which also shows that there is no significant difference in the NoS between the three algorithms in the case of $D=30$.

### 4.4.2 $D = 50$

The performance of updating velocities with LDS in 50 dimensional space is investigated in this subsection. The sizes of the exploration and exploitation subpopulations are $N_1=45$ and $N_2=75$, respectively. The maximum number of iterations is $G=10000$. Eleven test functions from Table 1 are used, and the other six test functions are discarded because all the considered algorithms are not able to find the average optimal solutions to these functions with relative errors smaller than $\varepsilon_{tol}=20\%$ in 10000 iterations. All other parameters involved are the same as those in Section 3. The CSs and the Friedman ranks of the different algorithms with different error tolerances are listed in Table 8. The NoSs of the different algorithms with different error tolerances are listed in Supplementary Table 2. The average convergence curves of the different algorithms with the test functions $F_3$ and $F_5$ are shown in Fig. 3, and the rest are shown in Supplementary Fig. 2.

Table 8 CSs and ranks of the three algorithms at different tolerance errors

| Fns. | $\varepsilon_{tol}=20\%$ | | | $\varepsilon_{tol}=10\%$ | | |
|---|---|---|---|---|---|---|
| | Rand | DES | OHS | Rand | DES | OHS |
| $F_1$ | 8106(1) | 8588(2) | 8664(3) | 8367(1) | 8871(2) | 8953(3) |
| $F_2$ | -(2.5) | 9010(2) | 8821(1) | -(2) | -(2) | -(2) |
| $F_3$ | 5734(3) | 2079(1) | 2139(2) | -(2) | -(2) | -(2) |
| $F_4$ | 25(3) | 23(2) | 18(1) | 563(3) | 261(2) | 259(1) |
| $F_5$ | -(3) | 2502(1) | 2650(2) | -(2) | -(2) | -(2) |
| $F_6$ | 3678(3) | 1764(2) | 1757(1) | -(2) | -(2) | -(2) |
| $F_7$ | 3805(3) | 1327(1) | 1330(2) | 4662(3) | 1492(2) | 1480(1) |
| $F_8$ | 3519(3) | 1579(1) | 1585(2) | -(3) | 2307(2) | 2288(1) |
| $F_{10}$ | -(2.5) | -(2.5) | 9797(1) | -(2) | -(2) | -(2) |
| $F_{13}$ | 7754(3) | 5886(2) | 5843(1) | -(2) | -(2) | -(2) |
| $F_{15}$ | 3445(3) | 1577(2) | 1534(1) | -(2) | -(2) | -(2) |
| AvgRks. | 2.773(3) | 1.682(2) | 1.546(1) | 2.227(3) | 2.046 (2) | 1.727(1) |

According to Table 8, the $\tau_F$ for $\varepsilon_{tol}=20\%$ and $10\%$ are 8.2642 and 0.6843, respectively. The Friedman critical value is $\tau_c=3.555$, and the Nemenyi critical difference is $CD=0.9995$. It is obvious that there is a significant difference among the three algorithms considered. When $\varepsilon_{tol}=20\%$, the velocity updating with OHS or DES significantly reduces the iterations required for the convergence. When $\varepsilon_{tol}=10\%$, the Nemenyi differences between the different algorithms are smaller than $CD$. This is mainly because there are only four test functions for which the considered algorithms can converge to $\varepsilon_{tol}=10\%$. If we perform the Nemenyi test on these four functions, DES and OHS perform much better than Rand. For the test function $F_8$, DES and OHS successfully find the optimal solution under the accuracy requirement $\varepsilon_{tol}=10\%$, but Rand fails. From the statistical results of the NoSs, it can be seen that there is no significant difference in the NoSs of the three algorithms. From the average convergence curves it can still be observed that updating velocities with OHS or DES can significantly improve the convergence speed of HCLPSO in 50 dimensional space.

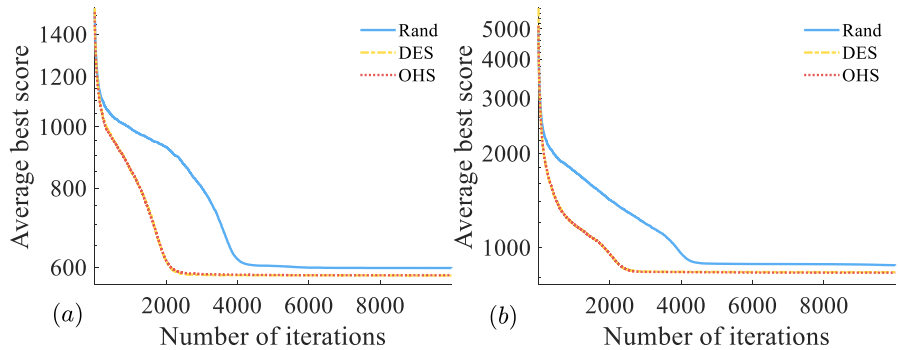

Fig 3 The average convergence curves of the three algorithms with: (a) $F_3$; and (b) $F_5$

## 5. Conclusion

The LDS covers the search space better than the random sequence, and hence is often used to improve the search ability of PSO. However, it is still not definite whether the use of LDS to generate the initial population can effectively improve the numerical performance of PSO. By using a modified PSO, namely HCLPSO, and four LDSs, i.e., HWS, OHS, DES and OA, this problem has been studied in detail and in depth.

The influence of generating the initial population with different LDSs was studied first. The numerical results show that the initial population generated by LDS can produce a degree of enhancement in the early iterations of HCLPSO but cannot greatly improve the numerical performance in the entire iterative process. For this observation, we explain that the inherent randomness in HCLPSO will iteratively weaken the enhancement of the early search ability brought by the LDS initial population as the iteration proceeds. If this explanation is correct, a deduction can be naturally given: Appropriately reducing the randomness in the iteration formula of HCLPSO can improve the convergence speed and computation time. To verify this deduction and explanation, we further carried out corresponding numerical experiments which show the following:

1) If the two key random sequences, i.e., $\varepsilon_{g,1}$ and $\varepsilon_{g,2}$ in Eqs. (5) and (6), are replaced with the deterministic LDS, given the same error tolerance, the convergence speed of HCLPSO can be significantly improved.

2) Given the same error tolerance, replacing $\varepsilon_{g,1}$ and $\varepsilon_{g,2}$ with any type of LDS can improve the convergence speed of HCLPSO without decreasing its success rate.

3) Updating the velocity with LDS can improve the convergence speed of HCLPSO in both low and high dimensional space.

This paper seems to propose a very basic approach, which emphasizes the appropriate reduction of randomness in the iterative formula of HCLPSO/ and replaces the random sequence with the deterministic LDS. This approach may be extended to other PSO type algorithms, and even to other types of evolutionary algorithms. As we end of this paper, please allow us to present an open question: For general evolutionary algorithms, can the numerical performance be improved by appropriately reducing the randomness in its iterative formula with the help of LDS? We hope evolutionary algorithm researchers will be interested in this problem.

# supplementary materials

Supplementary Table I  NoSs and ranks of the three algorithms with different error tolerances

|  | $\varepsilon_{tol} = 20\%$ | | | $\varepsilon_{tol} = 10\%$ | | |
| --- | --- | --- | --- | --- | --- | --- |
|  | Rand | DES | OHS | Rand | DES | OHS |
| $F_1$ | 60(2) | 60(2) | 60(2) | 60(2) | 60(2) | 60(2) |
| $F_2$ | 43(3) | 53(1) | 51(2) | 3(3) | 20(1) | 13(2) |
| $F_3$ | 60(2) | 60(2) | 60(2) | 53(2) | 56(1) | 48(3) |
| $F_4$ | 60(2) | 60(2) | 60(2) | 60(2) | 60(2) | 60(2) |
| $F_5$ | 60(2) | 60(2) | 60(2) | 7(3) | 34(1) | 28(2) |
| $F_6$ | 60(2) | 60(2) | 60(2) | 60(2) | 60(2) | 60(2) |
| $F_7$ | 60(2) | 60(2) | 60(2) | 60(2) | 60(2) | 60(2) |
| $F_8$ | 60(2) | 60(2) | 60(2) | 59(2) | 60(1) | 58(3) |
| $F_{10}$ | 42(3) | 55(1.5) | 55(1.5) | 31(3) | 38(1) | 34(2) |
| $F_{12}$ | 60(2) | 60(2) | 60(2) | 43(2) | 49(1) | 36(3) |
| $F_{13}$ | 54(2) | 56(1) | 52(3) | 48(2) | 53(1) | 43(3) |
| $F_{14}$ | 60(2) | 60(2) | 60(2) | 52(1) | 50(2) | 38(3) |
| $F_{15}$ | 60(2) | 60(2) | 60(2) | 8(2) | 4(3) | 10(1) |
| $F_{16}$ | 60(2) | 60(2) | 60(2) | 60(2) | 60(2) | 60(2) |
| $F_{17}$ | 38(3) | 54(1) | 47(2) | 1(3) | 13(1) | 3(2) |
| Avgks. | 2.200(3 | 1.767(1 | 2.033(2 | 2.267(2 | 1.755(3 | 1.533(1 |

Supplementary Table II  NoSs and ranks of the three algorithms with different error tolerances

|  | $\varepsilon_{tol} = 20\%$ | | | $\varepsilon_{tol} = 10\%$ | | |
|---|---|---|---|---|---|---|
|  | Rand | DES | OHS | Rand | DES | OHS |
| $F_1$ | 60(2) | 60(2) | 60(2) | 60(2) | 60(2) | 60(2) |
| $F_2$ | 38(3) | 43(2) | 44(1) | 29(1) | 24(3) | 27(2) |
| $F_3$ | 33(3) | 53(1) | 50(2) | 0(3) | 2(1) | 1(2) |
| $F_4$ | 60(2) | 60(2) | 60(2) | 60(2) | 60(2) | 60(2) |
| $F_5$ | 5(3) | 47(1) | 41(2) | 0(2) | 0(2) | 0(2) |
| $F_6$ | 60(2) | 60(2) | 60(2) | 20(3) | 33(2) | 36(1) |
| $F_7$ | 54(3) | 60(1.5) | 60(1.5) | 48(3) | 60(1.5) | 60(1.5) |
| $F_8$ | 59(3) | 60(1.5) | 60(1.5) | 22(3) | 45(2) | 46(1) |
| $F_{10}$ | 57(1) | 34(3) | 46(2) | 22(1) | 6(3) | 10(2) |
| $F_{13}$ | 55(1) | 52(3) | 53(2) | 47(1) | 37(3) | 45(2) |
| $F_{15}$ | 60(2) | 60(2) | 60(2) | 0(2) | 0(2) | 0(2) |
| AvgRks | 2.273(3 | 1.909(2 | 1.818(1 | 2.091(3 | 2.136 | 1.772(1 |

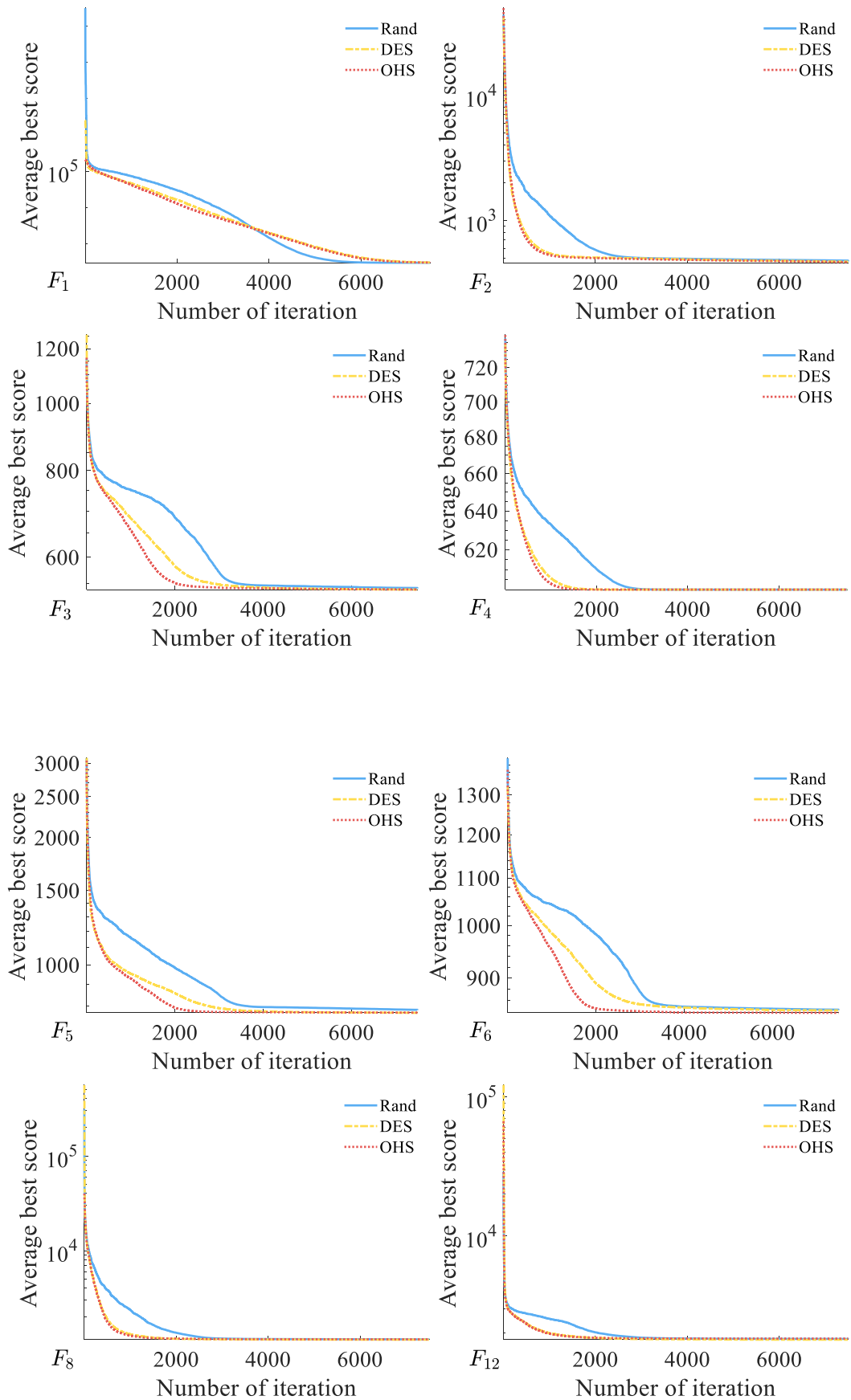

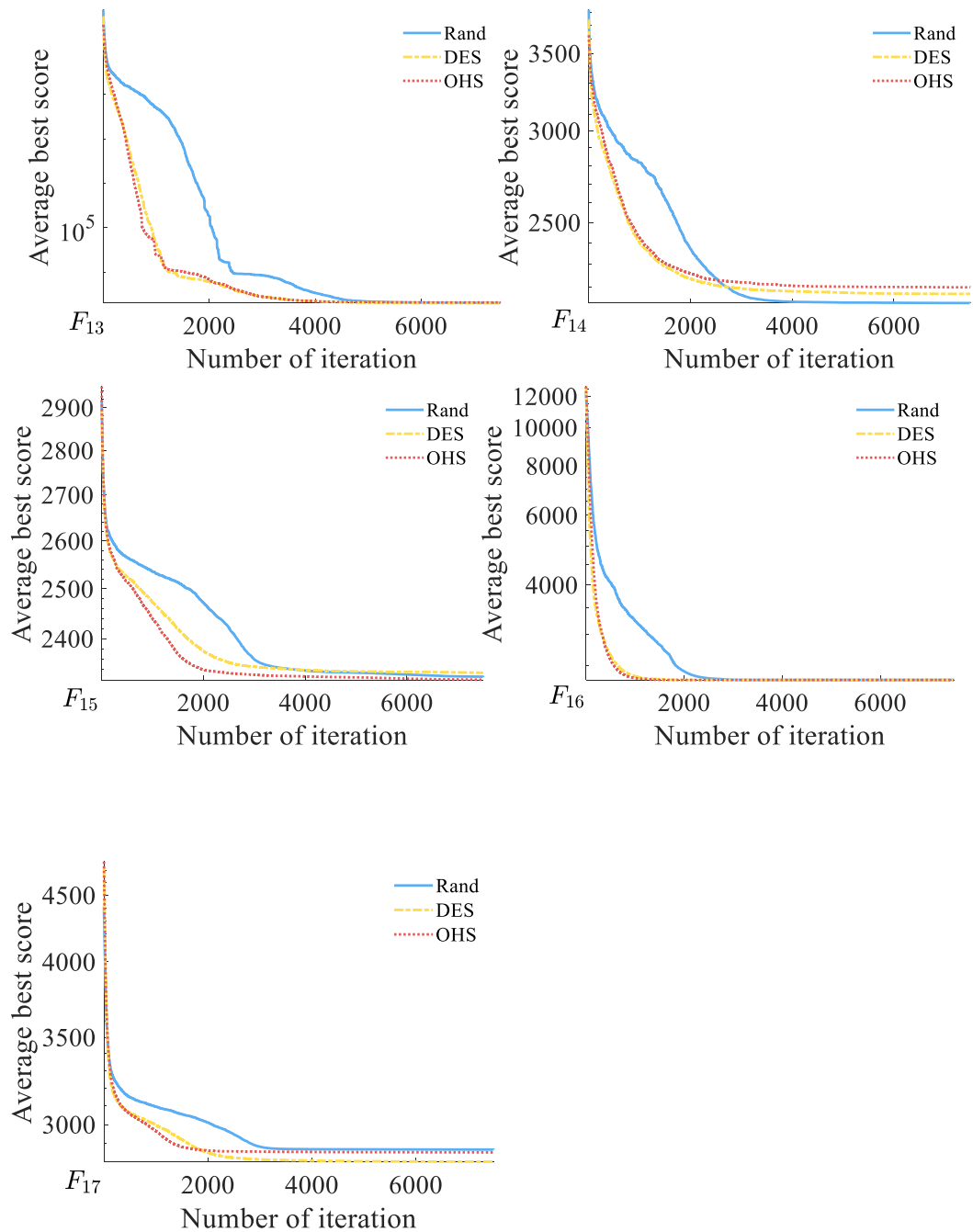

Supplementary Fig. 1 The average convergence curves of the three algorithms in $D = 30$

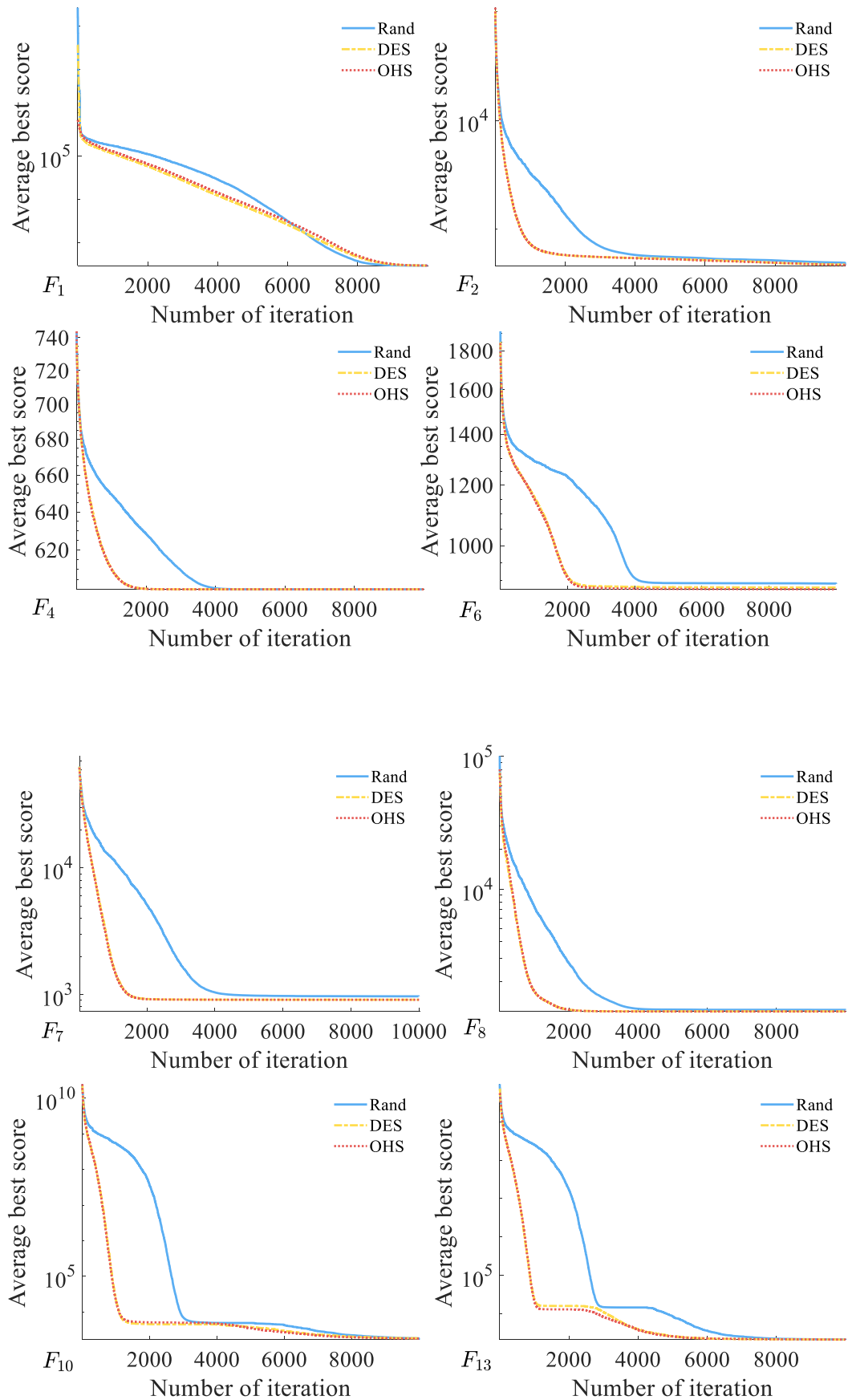

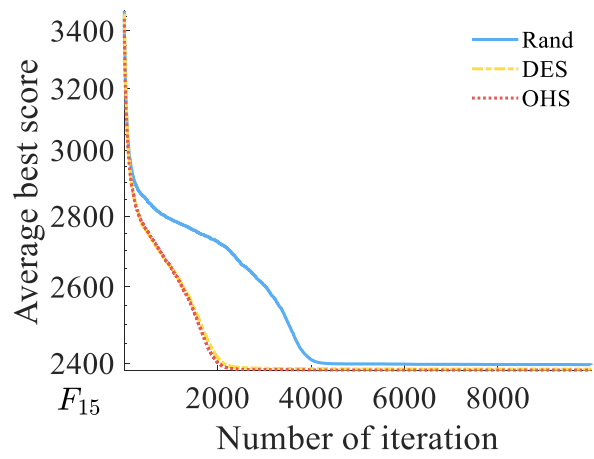

Supplementary Fig. 2 The average convergence curves of the three algorithms in $D = 50$